\documentclass{article}
\usepackage{amsmath,amsthm,amssymb,graphicx,url,fullpage}

\graphicspath{{./Fig/}}

\def\st{\ :\ }
\def\JB{\mathrm{JB}}
\def\JS{\mathrm{JS}}

\def\dmu{\mathrm{d}\mu}
\def\bbR{\mathbb{R}}

\def\calF{\mathcal{F}}

\def\calT{\mathcal{T}}
\def\calB{\mathcal{B}}
\def\calJ{\mathcal{J}}

\def\eqdef{:=}

\def\IS{\mathrm{IS}}
\def\KL{\mathrm{KL}}
\def\eKL{\mathrm{eKL}}

\def\JB{\mathrm{JB}}

\def\calO{\mathcal{O}}
\def\calC{\mathcal{C}}

\def\calX{\mathcal{X}}

\def\bbR{\mathbb{R}}

\title{The Bregman chord divergence}

\author{
  Frank Nielsen\\
	Sony Computer Science Laboratories Inc, Japan\\ 
	{\tt Frank.Nielsen@acm.org}
	\and
  Richard Nock\\
	Data 61, Australia\\ 
 The Australian National \& Sydney Universities \\ 
 {\tt Richard.Nock@data61.csiro.au}
}

\date{}

\begin{document}
 
\maketitle
 
\begin{abstract}
Distances are fundamental primitives whose choice significantly impacts the performances of algorithms in machine learning and signal processing.
However selecting the most appropriate distance for a given task is an endeavor.
Instead of testing one by one the entries of an ever-expanding dictionary of {\em ad hoc} distances, one rather prefers to consider parametric classes of distances that are exhaustively characterized by axioms derived from first principles.
Bregman divergences are such a class.
However fine-tuning a Bregman divergence is delicate since it requires to smoothly adjust a functional generator. 
In this work, we propose an extension of Bregman divergences called the Bregman chord divergences.
This new class of distances does not require gradient calculations, uses two scalar parameters that can be easily tailored in applications, and generalizes asymptotically Bregman divergences.
\end{abstract}

\section{Introduction}\label{sec:intro}
Dissimilarities (or distances) are at the heart of many signal processing tasks~\cite{Deza-2009,Basseville-2013}, 
and the performance of algorithms solving those tasks heavily depends on the chosen distances.
A {\em dissimilarity} $D(O_1:O_2)$ between two objects $O_1$ and $O_2$ belonging to a space $\calO$ (e.g., vectors, matrices, probability densities, random variables, etc.) is a function $D: \calO\times\calO\rightarrow [0,+\infty]$ 
such that $D(O_1:O_2)\geq 0$ with equality if and only if $O_1=O_2$.
Since a dissimilarity may not be symmetric (i.e., an oriented dissimilarity with $D(O_1:O_2)\not=D(O_2:O_1)$), we emphasize this
fact using the notation\footnote{In information theory~\cite{CT-IT-2012}, the double bar  notation '$||$' has been used to avoid confusion with the comma ',' notation, used for example in joint entropy $H(X,Y)$.} ':'. The {\em reverse} dissimilarity or {\em dual} dissimilarity is defined by 
\begin{equation}
D^*(O_1:O_2) \eqdef D(O_2:O_1),
\end{equation}
and satisfies the involutive property: $({D^*})^*=D$.
When a symmetric dissimilarity further satisfies the triangular inequality 
\begin{equation}
D(O_1,O_2)+D(O_2,O_3)\geq D(O_1,O_3), \quad\forall O_1, O_2, O_3\in\calO,
\end{equation}
it is called a {\em metric distance}. 

Historically, many {\it ad hoc} distances have been proposed and empirically benchmarked on different tasks in order to improve the state-of-the-art performances. However, getting the most appropriate distance for a given task is often an endeavour.
Thus principled classes of distances\footnote{Here, we use the word distance to mean a dissimilarity (or a distortion), not necessarily a metric distance~\cite{Deza-2009}. A distance satisfies $D(\theta_1,\theta_2)\geq 0$ with equality iff. $\theta_1=\theta_2$.} have been proposed and studied.
Among those generic classes of distances, three main types have emerged:

\begin{itemize}
	\item The {\em Bregman divergences}~\cite{Bregman-1967,BregmanKmeans-2005} defined for a strictly convex and differentiable generator $F\in\calB:\Theta\rightarrow\bbR$:
	\begin{equation}\label{eq:BD}
B_F(\theta_1:\theta_2) \eqdef F(\theta_1)-F(\theta_2)-(\theta_1-\theta_2)^\top \nabla F(\theta_2),
\end{equation} 
measure the dissimilarity between {\em parameters} $\theta_1,\theta_2\in\Theta$.
We use the term ``divergence'' (rooted in information geometry~\cite{IG-2016}) instead of distance to emphasize the smoothness property\footnote{A metric distance is not smooth at its calling arguments.} of the distance.
The dual Bregman divergence $B_F^*(\theta_1:\theta_2)$ is obtained from the Bregman divergence induced by the 
Legendre convex conjugate:
 
\begin{equation}
B_F^*(\theta_1:\theta_2) \eqdef B_F(\theta_2:\theta_1) = B_{F^*}(\nabla F(\theta_1):\nabla F(\theta_2)),
\end{equation} 
where the Legendre-Fenchel transformation is defined by:
\begin{equation}
F^*(\eta)=\sup_{\theta\in\Theta} \{\theta^\top\eta-F(\theta)\}.
\end{equation}

\item The Csisz\'ar $f$-divergences~\cite{AliSilvey-1966,Csiszar-1967} defined for a convex generator $f\in\calC$ satisfying $f(1)=0$:
\begin{equation}
I_f[p_1:p_2] \eqdef \int_\calX p_1(x) f\left(\frac{p_2(x)}{p_1(x)}\right) \dmu(x),
\end{equation} 
measure  the dissimilarity between {\em probability densities} $p$ and $q$ that are absolutely continuous with respect to a base measure $\mu$ (defined on a support $\calX$).
A {\em scalar divergence} is a divergence acting on scalar parameters, i. e., a 1D divergence.
A {\em separable divergence} is a divergence that can be written as a sum of elementary scalar divergences.
The $f$-divergences are separable divergences since we have: 
\begin{equation}
I_f[p:q] = \int i_f[p(x):q(x)]  \dmu(x),
\end{equation}
with the scalar $f$-divergence $i_f[a:b]\eqdef a f\left(\frac{b}{a}\right)$.

The dual $f$-divergence is obtained for the generator $f^\diamond(u)\eqdef uf(1/u)$ (diamond $f$-generator) as follows:
\begin{equation}
I_f^*[p:q] \eqdef  I_f[q:p]= I_{f^\diamond}[p:q].
\end{equation} 
We may $J$-symmetrize\footnote{By analogy to the Jeffreys divergence that is the symmetrized Kullback-Leibler divergence.} a $f$-divergence
	by defining its generator $f^\circ$:
\begin{eqnarray}
J_{f}[p:q] &=& \frac{1}{2} (I_f[p:q]+I_f[q:p]),\\
&=& I_{f^\circ}[p,q],
\end{eqnarray}
with
$$
f^\circ(u)   \eqdef \frac{1}{2}\left(f(u)+f^*(u)\right).
$$
Alternatively, we may JS-symmetrize\footnote{By analogy to the Jensen-Shannon divergence (JS).}  the $f$-divergence 
by using the following  generator $f^\bullet$:
\begin{eqnarray}
\JS_{f}[p:q] &:=& \frac{1}{2} \left(I_f\left[p:\frac{p+q}{2}\right]+I_f\left[q:\frac{p+q}{2}\right]\right),\\
&=& I_{f^\bullet}[p,q],\\
f^\bullet(u)&\eqdef& \frac{1+u}{4} \left(f\left(\frac{2u}{1+u}\right) +  f\left(\frac{2}{1+u}\right)  \right).
\end{eqnarray}

\item The Burbea-Rao divergences~\cite{BurbeaRao-1982} also called Jensen divergences because they rely on the Jensen's inequality~\cite{Jensen-1906} for a strictly convex function $F\in\calJ:\Theta\rightarrow\bbR$:
\begin{equation}\label{eq:JD}
J_F(\theta_1,\theta_2) \eqdef \frac{F(\theta_1)+F(\theta_2)}{2}  -F \left(\frac{\theta_1+\theta_2}{2}\right) \geq 0.
\end{equation}
We note in passing that Bregman divergences can be extended to strictly convex and non-differentiable generator as well~\cite{GenBregman-1997,AggloBD-2012}.
\end{itemize}

These  three fundamental classes of distances are {\em not} mutually exclusive, and their pairwise intersections (e.g., $\calB\cap\calC$ or $\calJ\cap\calC$) have been studied in~\cite{Pardo-1997,Amari-2009,InformationMonotonicityBinary-2014}.
The ':' notation between arguments of distances emphasizes the potential asymmetry of distances 
(oriented distances with $D(\theta_1:\theta_2)\not = D(\theta_2:\theta_1)$), and the brackets surrounding distance arguments indicate that it is a {\em statistical distance} between probability densities, and not a distance between parameters.
Using these notations, we express the Kullback-Leibler distance~\cite{CT-IT-2012} (KL)  as: 
\begin{equation}
\KL[p_1:p_2] \eqdef \int p_{1}(x)\log \frac{p_{1}(x)}{p_{2}(x)}\dmu(x).
\end{equation} 
The KL distance between two members $p_{\theta_1}$ and $p_{\theta_2}$ of a parametric family $\calF$ of distributions amount to a parameter divergence:
\begin{equation}
\KL_\calF(\theta_1:\theta_2) \eqdef \KL[p_{\theta_1}:p_{\theta_2}].
\end{equation}
For example, the KL statistical distance between two probability densities belonging to the same exponential family or the same mixture family amounts to a (parameter) Bregman divergence~\cite{IG-2016,geowmixtures-2018}.
When $p_1$ and $p_2$ are finite discrete distributions of the $d$-dimensional probability simplex $\Delta_d$, we have
$\KL_{\Delta_d}(p_1:p_2)=\KL[p_{1}:p_{2}]$. 
This explains why sometimes we can handle loosely distances between discrete distributions as both a parameter distance and a statistical distance.
For example, the KL distance between two discrete distributions is a Bregman divergence $B_{F_\KL}$ for $F_\KL(x)=\sum_{i=1}^d x_i\log x_i$ (Shannon negentropy) for $x\in \Theta=\Delta_d$.
Extending $\Theta=\Delta_d$ to positive measures $\Theta=\bbR_+^d$, this Bregman divergence $B_{F_\KL}$ yields the extended KL distance:
$\eKL[p:q] = \sum_{i=1}^d p_{i}\log \frac{p_{i}}{q_{i}}+q_i-p_i$.

Whenever using a functionally parameterized distance in applications, we need to choose the most appropriate functional generator, ideally from first principles~\cite{Csiszar-1991,Banerjee-2005,IG-2016}.
For example, Non-negative Matrix Factorization (NMF) for audio source separation
or music transcription from the signal power spectrogram
can be done by selecting the Itakura-Saito divergence~\cite{Fevotte-2011} (a Bregman divergence for the Burg negentropy $F_\IS(x)=-\sum_i\log x_i$) that satisfies the requirement of being {\em scale invariant}: $B_{F_\IS}(\lambda\theta:\lambda\theta')=B_{F_\IS}(\theta:\theta')=\sum_i \frac{\theta_i}{\theta_i'}-\log \frac{\theta_i}{\theta_i'}-1$ for any $\lambda>0$.  
When no such first principles can be easily stated for a task~\cite{Csiszar-1991}, we are left by choosing manually or by cross-validation a generator.
Notice that the convex combinations of Csisz\'ar generators is a Csisz\'ar generator (idem for Bregman divergences):
$\sum_i \lambda_i I_{f_i}=I_{\sum_i \lambda_i f_i}$ for $\lambda$ belonging to the standard simplex $\Delta_d$. 
Thus in practice, we could choose a base of generators and learn the best distance weighting (by analogy to feature weighting~\cite{Modha-2003}).
However, in doing so, we are left with the problem of choosing the base generators, and moreover we need to sum up different distances:
 This raises the problem of properly adding distance units! 
Thus in applications, it is often preferable to consider a smooth family of generators parameterized by scalars (e.g., $\alpha$-divergences~\cite{alphadivNMF-2008} or $\beta$-divergences~\cite{betablind-2002}, etc), and then finely tune these scalars.

In this work, we propose a novel class of distances, termed Bregman chord divergences.
Bregman chord divergences are parameterized by two scalar parameters which make it easy to fine-tune in applications, and matches asymptotically the ordinary Bregman divergences. 

The paper is organized as follows: 
In \S\ref{sec:skew}, we describe the skewed Jensen divergence, show how to biskew any distance by using two scalars, and report on the  Jensen chord divergence.
In \S\ref{sec:chordBregman1D}, we first introduce the univariate Bregman chord divergence, and then extend its definition to the multivariate case, in \S\ref{sec:chordBregmandD}. Finally, we conclude in~\S\ref{sec:concl}.

\section{Geometric design of skewed divergences from graph plots}\label{sec:skew}

We can geometrically {\em design}  divergences from convexity gap properties of the plot of the generator.
For example, the Jensen divergence $J_F(\theta_1:\theta_2)$ of Eq.~\ref{eq:JD} is visualized as the ordinate (vertical) gap between the midpoint of the line segment  
$[(\theta_1,F(\theta_1));(\theta_2,F(\theta_2))]$ and the point $(\frac{\theta_1+\theta_2}{2},F(\frac{\theta_1+\theta_2}{2}))$.
The non-negativity property of the Jensen divergence follows from the Jensen's midpoint convex inequality~\cite{Jensen-1906}.
Instead of taking the midpoint  $\bar\theta=\frac{\theta_1+\theta_2}{2}$, we can take {\em any} interior point 
$(\theta_1\theta_2)_\alpha \eqdef (1-\alpha)\theta_1+\alpha\theta_2$, and get the skewed $\alpha$-Jensen divergence (for any $\alpha\in (0,1)$):
\begin{equation}
J_F^\alpha(\theta_1:\theta_2) \eqdef  (F(\theta_1)F(\theta_2))_\alpha  - F((\theta_1\theta_2)_\alpha) \geq 0.
\end{equation}
A remarkable fact is that the scaled $\alpha$-Jensen divergence $\frac{1}{\alpha}J_F^\alpha(\theta_1:\theta_2)$ tends asymptotically to the reverse Bregman divergence $B_F(\theta_2:\theta_1)$ when $\alpha\rightarrow 0$, see~\cite{Zhang-2004,JD-2011}.
Notice that the Jensen divergences can be interpreted as Jensen-Shannon-type symmetrization~\cite{JB-2011} of Bregman divergences:
\begin{equation}
J_F(\theta_1:\theta_2)=   B_F\left(\theta_1:\frac{\theta_1+\theta_2}{2}\right) + B_F\left(\theta_2:\frac{\theta_1+\theta_2}{2}\right),
\end{equation}
and more generally, we have the skewed Jensen-Bregman divergences:
\begin{equation}
\JB_F^\alpha(\theta:\theta')  \eqdef (1-\alpha)B_F(\theta:(\theta\theta')_\alpha)+\alpha B_F(\theta':(\theta\theta')_\alpha).
\end{equation}

By measuring the ordinate gap between two non-crossing upper and lower chords anchored at the generator graph plot, we can extend the $\alpha$-Jensen divergences to a tri-parametric family of Jensen chord divergences~\cite{chordgap-2018}:
\begin{equation}
J_F^{\alpha,\beta,\gamma}(\theta:\theta') \eqdef (F(\theta)F(\theta'))_\gamma -(F((\theta\theta')_\alpha)F((\theta\theta')_\beta))_{\frac{\gamma-\alpha}{\beta-\alpha}},
\end{equation}
with $\alpha,\beta\in [0,1]$ and $\gamma\in [\alpha,\beta]$. The $\alpha$-Jensen divergence is recovered when $\alpha=\beta=\gamma$.

\begin{figure}
\centering
\includegraphics[width=0.6\columnwidth]{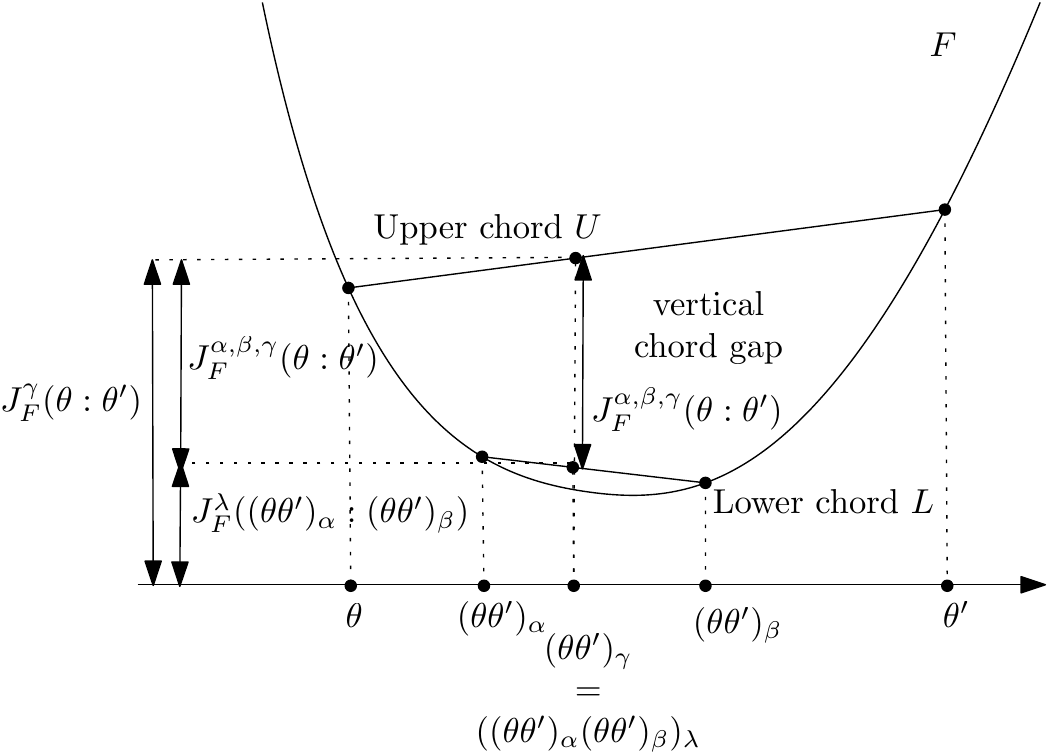}

\caption{The Jensen chord gap divergence. \label{fig:JensenChord}}
\end{figure}

For any given distance $D:\Theta\times\Theta\rightarrow\bbR_+$ (with convex parameter space $\Theta$), we can biskew the distance by considering two scalars $\gamma,\delta\in\bbR$ (with $\delta\not=\gamma$) as:
\begin{equation}\label{eq:biskew}
D_{\gamma,\delta}(\theta_1:\theta_2) \eqdef  D((\theta_1\theta_2)_\gamma:(\theta_1\theta_2)_\delta).
\end{equation}
Clearly, $(\theta_1\theta_2)_\gamma=(\theta_1\theta_2)_\delta$ iff.  $(\delta-\gamma)(\theta_1-\theta_2)=0$.
That is, if (i) $\theta_1=\theta_2$ or if (ii) $\delta=\gamma$. Since by definition $\delta\not=\gamma$, 
we have $D_{\gamma,\delta}(\theta_1:\theta_2)=0$ iff $\theta_1=\theta_2$.
Notice that both $(\theta_1\theta_2)_\gamma=(1-\gamma)\theta_1+\gamma\theta_2$ and $(\theta_1\theta_2)_\delta=(1-\delta)\theta_1+\delta\theta_2$ should belong to the parameter space $\Theta$. A sufficient condition is to ensure that $\gamma,\delta\in [0,1]$ so that both
$(\theta_1\theta_2)_\gamma\in\Theta$ and $(\theta_1\theta_2)_\delta\in\Theta$.
When $\Theta=\bbR^d$, we may further consider any $\gamma,\delta\in\bbR$.

\section{The scalar Bregman chord divergence}\label{sec:chordBregman1D}

\begin{figure}
	\centering
	\includegraphics[width=0.6\textwidth]{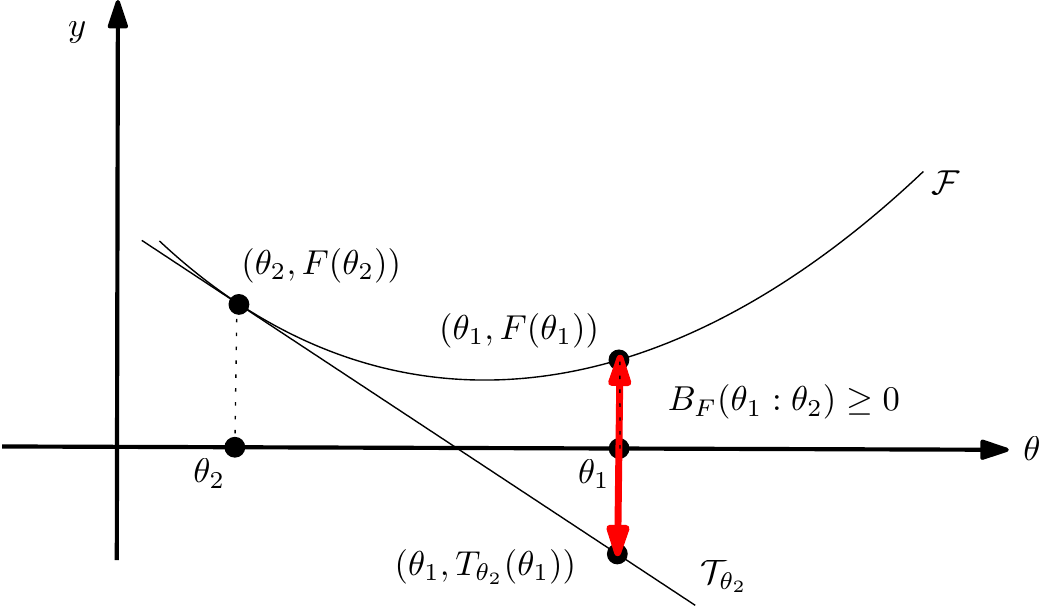}
	
	\caption{Illustration of the univariate Bregman divergence as the ordinate gap (`vertical' gap) evaluated at $\theta_1$ between the graph plot $\calF$ and the tangent line $\calT_{\theta_2}$ to $\calF$ at $\theta_2$.
	\label{fig:BD1D}}
\end{figure}

Let $F:\Theta\subset\bbR\rightarrow\bbR$ be a  univariate Bregman generator with open convex domain $\Theta$, and
denote by $\calF=\{ (\theta,F(\theta)) \}_\theta$ its  graph. 
Let us rewrite the ordinary univariate Bregman divergence~\cite{Bregman-1967} of Eq.~\ref{eq:BD} as follows:
\begin{equation}\label{eq:BD1D}
B_F(\theta_1:\theta_2) =  F(\theta_1) - T_{\theta_2}(\theta_1),   
\end{equation}
where $y=T_{\theta}(\omega)$ denotes the equation of the tangent line of $F$ at $\theta$:
\begin{equation}\label{eq:TL1D}
T_{\theta}(\omega) \eqdef F(\theta)+(\omega-\theta) F'(\theta),  
\end{equation}
Let $\calT_\theta=\{ (\theta,T_\theta(\omega)) \st \theta\in\Theta\}$ denote the graph of that tangent line.
Line $\calT_\theta$ is tangent to curve $\calF$ at point $P_\theta\eqdef (\theta,F(\theta))$.
Graphically speaking, the Bregman divergence is interpreted as the {\em ordinate  gap} (gap vertical) between the point $P_{\theta_1}=(\theta_1,F(\theta_1))\in\calF$ and 
the point of
$(\theta_1,T_{\theta_2}(\theta_1))\in\calT_\theta$, as depicted in Figure~\ref{fig:BD1D}.

\begin{figure}
	\centering
	\includegraphics[width=0.6\columnwidth]{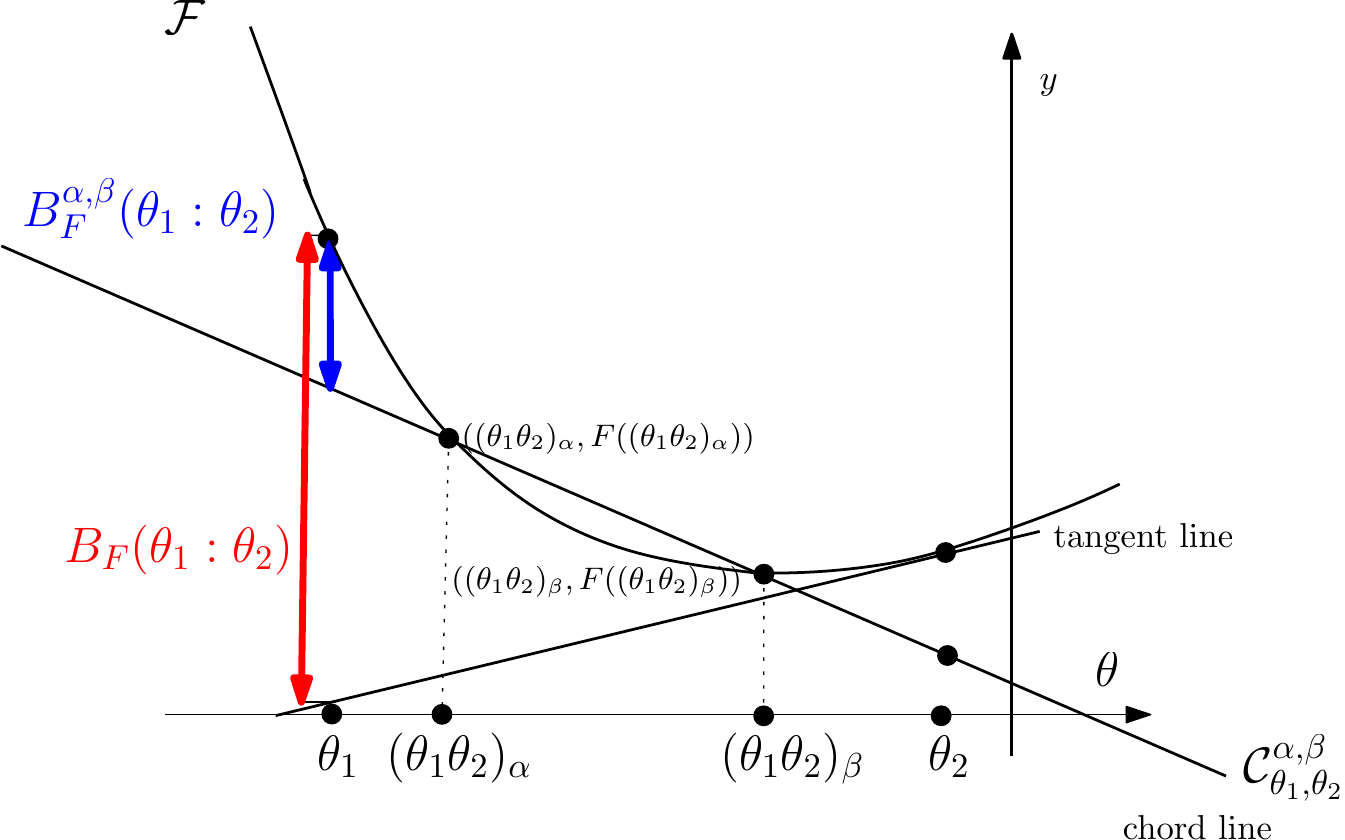}
	
	\caption{The Bregman chord  divergence $B_F^{\alpha,\beta}(\theta_1:\theta_2)$.
	\label{fig:CBD1D}}
\end{figure}

Now let us observe that we may {\em relax} the tangent line $\calT_{\theta_2}$ to a {\em chord line} (or secant)
 $\calC_{\theta_1,\theta_2}^{\alpha,\beta} = \calC_{(\theta_1\theta_2)_\alpha,(\theta_1\theta_2)_\beta}$  passing through the points $((\theta_1\theta_2)_\alpha,F((\theta_1\theta_2)_\alpha))$ and $((\theta_1\theta_2)_\beta,F((\theta_1\theta_2)_\beta))$
 for $\alpha,\beta\in (0,1)$ with $\alpha\not=\beta$ (with corresponding Cartesian equation $C_{(\theta_1\theta_2)_\alpha,(\theta_1\theta_2)_\beta}$), and still get a non-negative vertical gap between $(\theta_1,F(\theta_1))$ and $(\theta_1,C_{(\theta_1\theta_2)_\alpha,(\theta_1\theta_2)_\beta}(\theta_1))$ (because any line intersects a convex in at most two points).
By construction, this vertical gap is smaller than the gap measured by the ordinary Bregman divergence.
This yields the Bregman chord  divergence  ($\alpha,\beta\in (0,1]$, $\alpha\not=\beta$):
\begin{equation}
B_F^{\alpha,\beta}(\theta_1:\theta_2) \eqdef F(\theta_1) - C_F^{(\theta_1\theta_2)_\alpha,(\theta_1\theta_2)_\beta}(\theta_1) \leq 
B_F(\theta_1:\theta_2),
\end{equation} 
illustrated in Figure~\ref{fig:CBD1D}.
By expanding the chord equation and massaging the equation, we get the formula:
\begin{eqnarray*}
\lefteqn{B_F^{\alpha,\beta}(\theta_1:\theta_2) \eqdef }\nonumber\\
&& F(\theta_1) - \Delta_F^{\alpha,\beta}(\theta_1,\theta_2) (\theta_1-(\theta_1\theta_2)_\alpha) - F((\theta_1\theta_2)_\alpha),\\
&& F(\theta_1) - F\left((\theta_1\theta_2)_\alpha\right) +   \frac{\alpha \left\{{F\left((\theta_1\theta_2)_\alpha\right)-F\left((\theta_1\theta_2)_\beta\right)}\right\}}{\beta-\alpha},\nonumber
\end{eqnarray*}
where 
$$
\Delta_F^{\alpha,\beta}(\theta_1,\theta_2)\eqdef \frac{F((\theta_1\theta_2)_\alpha)-F((\theta_1\theta_2)_\beta)}{(\theta_1\theta_2)_\alpha-(\theta_1\theta_2)_\beta},
$$
 is the slope of the chord, and since $(\theta_1\theta_2)_\alpha-(\theta_1\theta_2)_\beta=(\beta-\alpha)(\theta_1-\theta_2)$
and $\theta_1-(\theta_1\theta_2)_\alpha=\alpha(\theta_1-\theta_2)$.

Notice the symmetry: 
$$
B_F^{\alpha,\beta}(\theta_1:\theta_2)=B_F^{\beta,\alpha}(\theta_1:\theta_2).
$$

We have asymptotically:
$$
\lim_{\alpha\rightarrow 1,\beta\rightarrow 1} B_F^{\alpha,\beta}(\theta_1:\theta_2) = B_F(\theta_1:\theta_2).
$$
When $\alpha\rightarrow\beta$, the Bregman chord divergences yields a subfamily of  {\em Bregman tangent divergences}: 
$B_F^{\alpha}(\theta_1:\theta_2)=\lim_{\beta\rightarrow\alpha} B_F^{\alpha,\beta}(\theta_1:\theta_2)\leq B_F(\theta_1:\theta_2)$.
	We consider the tangent line $\calT_{(\theta_1\theta_2)_\alpha}$ at $(\theta_1\theta_2)_\alpha$ and measure the ordinate gap at 
	$\theta_1$ between the function plot and this tangent line:
	
	\begin{eqnarray}
B_F^\alpha(\theta_1:\theta_2) &\eqdef& F(\theta_1)-F\left((\theta_1\theta_2)_\alpha\right)-
	\left(\theta_1-(\theta_1\theta_2)_\alpha\right)^\top\nabla F\left((\theta_1\theta_2)_\alpha\right),\nonumber\\
	&=& F(\theta_1)-F\left((\theta_1\theta_2)_\alpha\right)-\alpha(\theta_1-\theta_2)^\top \nabla F\left((\theta_1\theta_2)_\alpha\right),
	\end{eqnarray}
  for $\alpha\in (0,1]$. 
	The ordinary Bregman divergence is recovered when $\alpha=1$.
	Notice that the {\em mean value theorem} yields $\Delta_F^{\alpha,\beta}(\theta_1,\theta_2)=F'(\xi)$ for $\xi\in (\theta_1,\theta_2)$.
  Thus $B_F^{\alpha,\beta}(\theta_1:\theta_2)=B_F^{\xi}(\theta_1:\theta_2)$ for $\xi \in (\theta_1,\theta_2)$.
	Letting $\beta=1$ and $\alpha=1-\epsilon$ (for small values of $1>\epsilon>0$), we can approximate the ordinary Bregman divergence by the Bregman chord divergence without requiring to compute the gradient:
	$B_F(\theta_1:\theta_2) \simeq_{\epsilon\rightarrow 0} B_F^{1-\epsilon,1}(\theta_1:\theta_2)$.

\begin{figure}
	\centering
	
	\begin{tabular}{|c|c|c|}
	& Bregman chord  divergence &  Bregman tangent divergence\\ \hline\hline
	Quadratic & \includegraphics[width=0.35\textwidth]{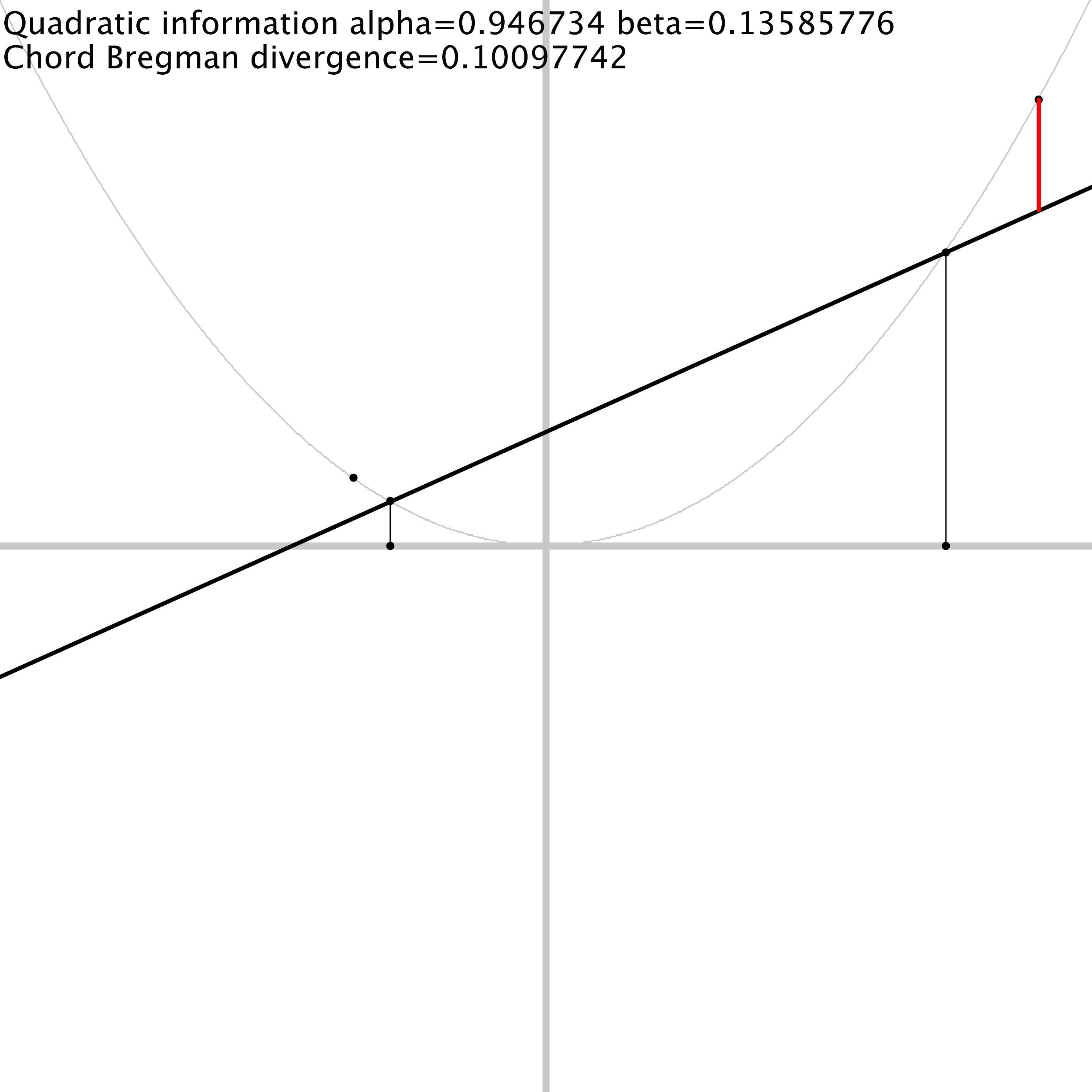} &
	\includegraphics[width=0.45\textwidth]{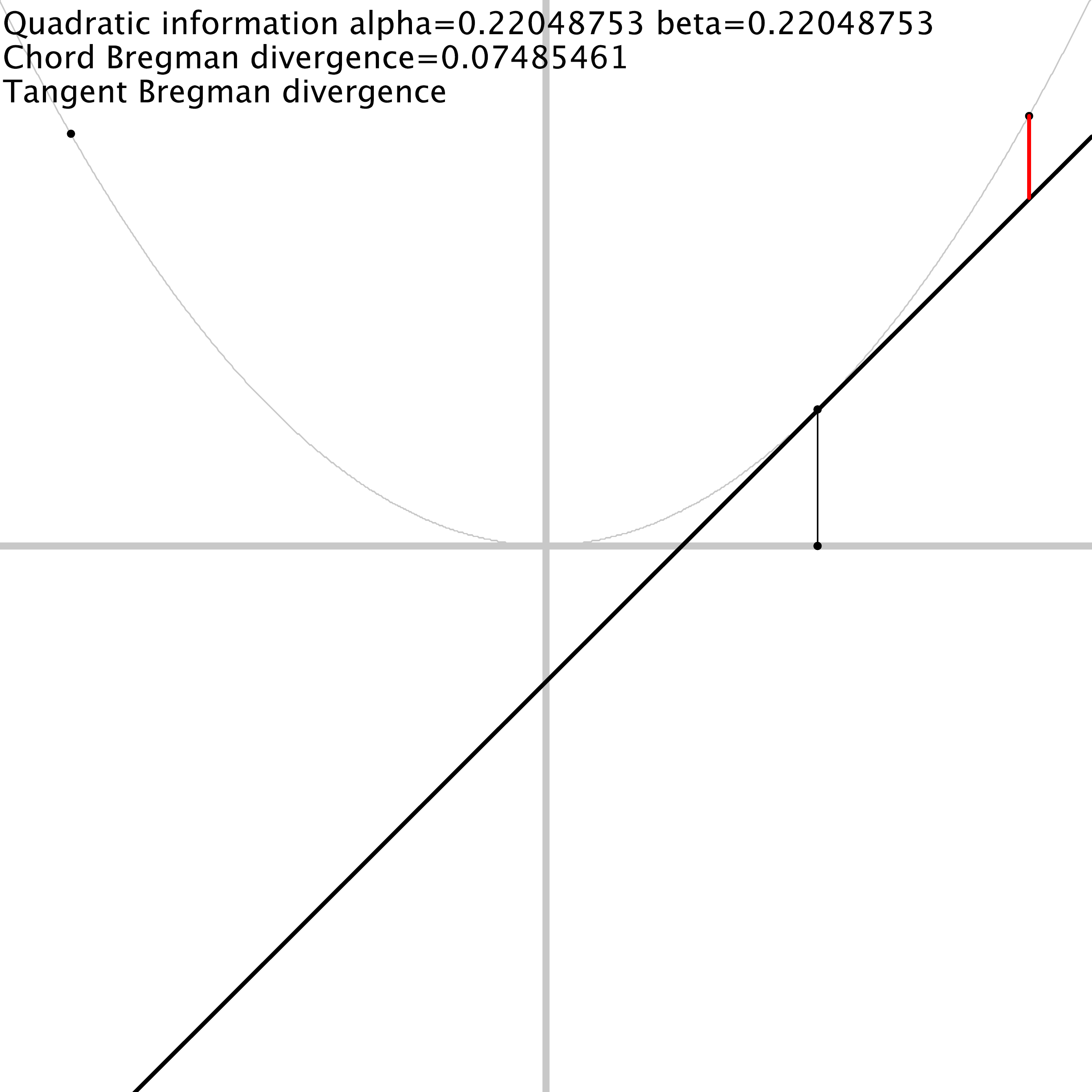} \\ \hline
	& (a) & (b)  \\ \hline
	Shannon & \includegraphics[width=0.35\textwidth]{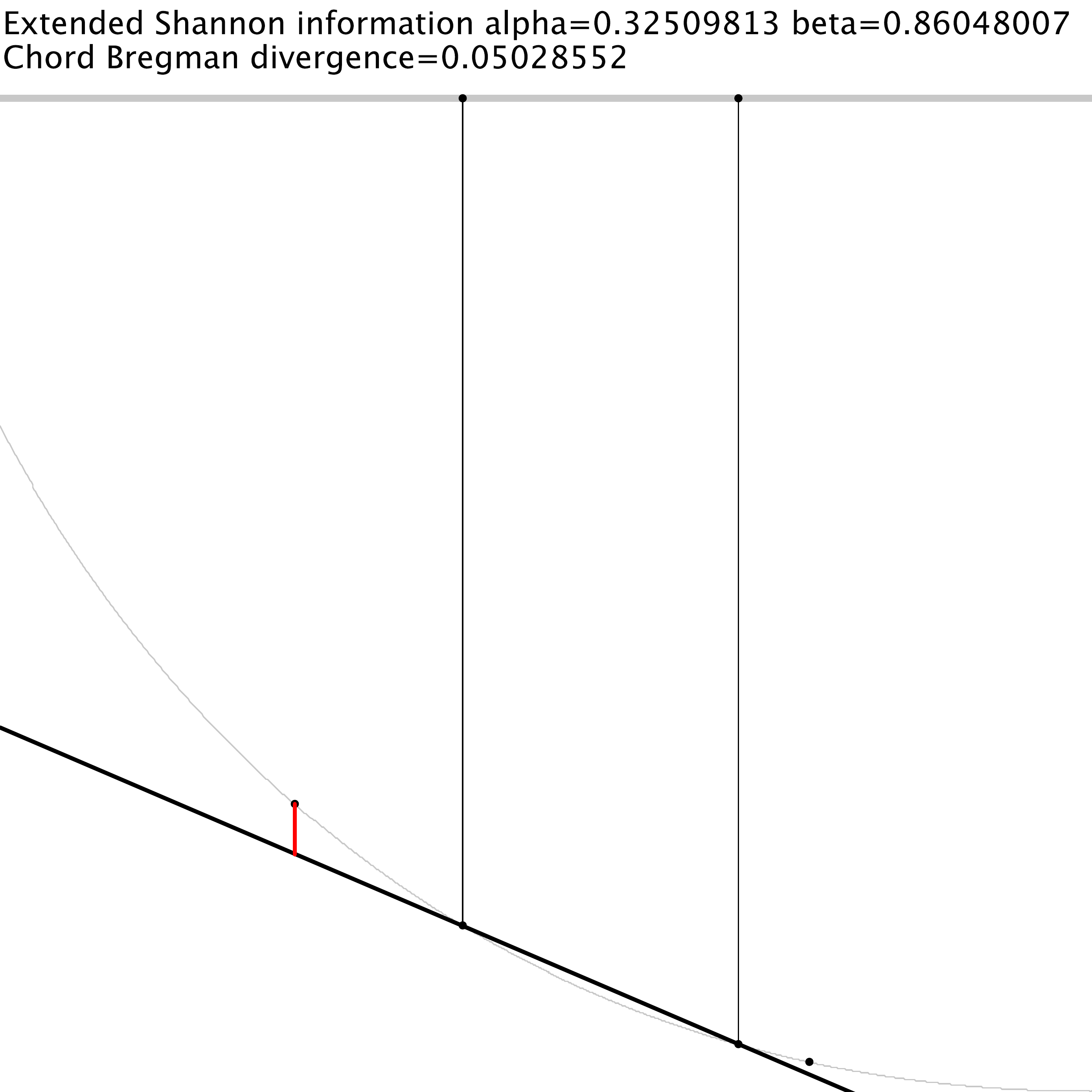} &
	\includegraphics[width=0.45\textwidth]{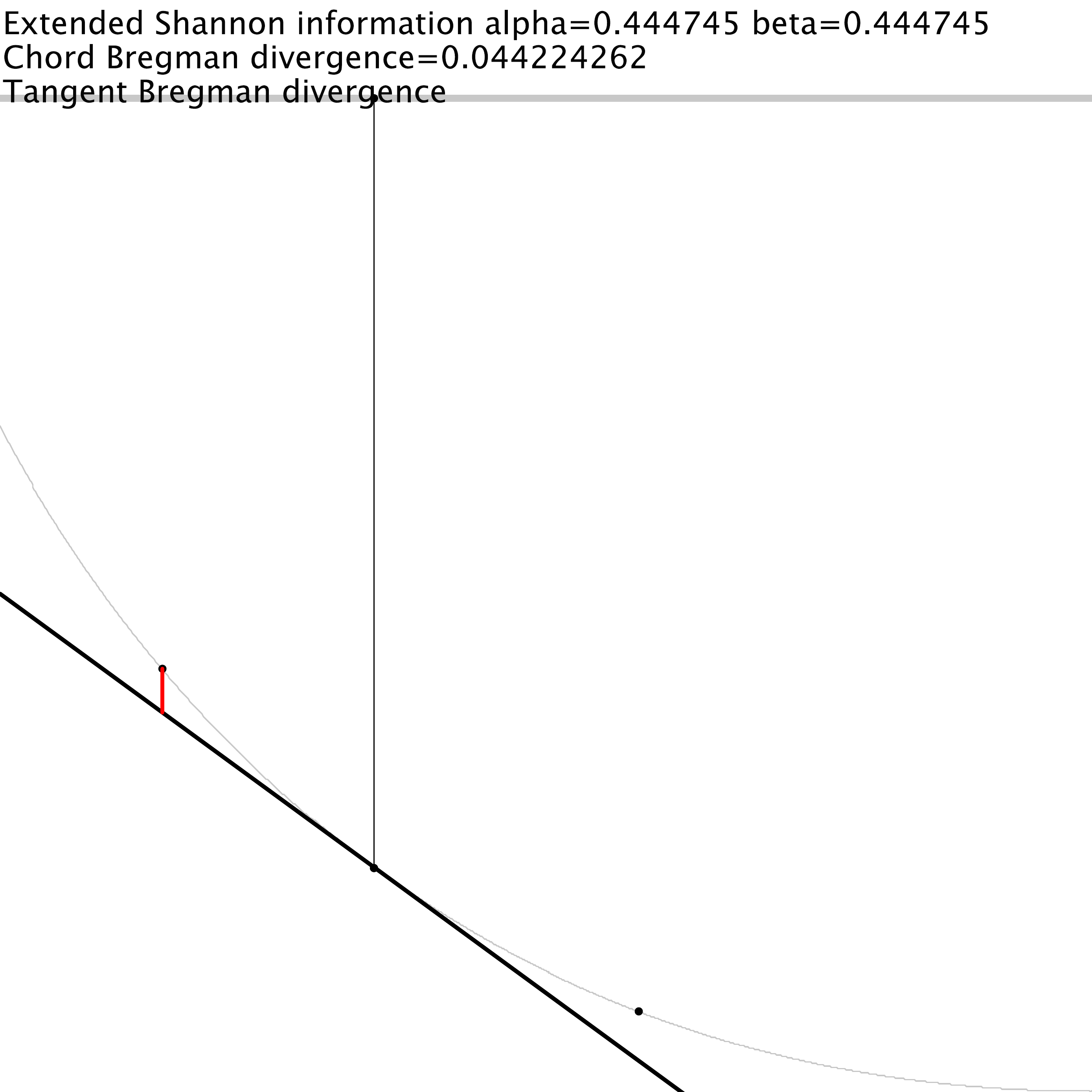} \\ \hline
	& (c) & (d)  \\ \hline
	\end{tabular}
	
	\caption{The univariate  Bregman chord divergences and  Bregman tangent divergences for the   quadratic and  Shannon information generators.
	\label{fig:chordBD}}
\end{figure}

 Figure~\ref{fig:chordBD} displays some snapshots of an interactive demo program that illustrates the impact of $\alpha$ and $\beta$ for defining the Bregman chord divergences for the quadratic and Shannon generators.

\section{The multivariate  Bregman chord divergence}\label{sec:chordBregmandD}
When the generator is separable~\cite{IG-2016}, i.e., $F(x)=\sum_i F_i(x_i)$ for univariate generators $F_i$, we extend easily
the Bregman chord divergence as: $B_F^{\alpha,\beta}(\theta:\theta')=\sum_i B_{F_i}^{\alpha,\beta}(\theta_i:\theta'_i)$. Otherwise, we have to carefully define the notion of ``slope'' for the multivariate case.
An example of such a non-separable multivariate generator is the Legendre dual of the Shannon negentropy: The log-sum-exp function $F(\theta)=\log(1+\sum_i e^{\theta_i})$.

Given a multivariate (non-separable) Bregman generator $F(\theta)$ with $\Theta\subseteq\bbR^D$ and  two prescribed distinct parameters $\theta_1$ and $\theta_2$,
 consider the following univariate function, for $\lambda\in\bbR$:
\begin{equation}
F_{\theta_1,\theta_2}(\lambda) \eqdef F\left((1-\lambda)\theta_1+\lambda\theta_2\right) = F\left(\theta_1+\lambda(\theta_2-\theta_1)\right),
\end{equation}
with $F_{\theta_1,\theta_2}(0)=F(\theta_1)$ and $F_{\theta_1,\theta_2}(1)=F(\theta_2)$.

The functions $\{F_{\theta_1,\theta_2}\}$ are strictly convex and univariate Bregman generators. 

\begin{proof}
To prove the strict convexity of a univariate function $G$, we need to show that for any $\alpha\in (0,1)$, we have
$$
G((1-\alpha)x+\alpha y)< (1-\alpha)G(x)+\alpha G(y).
$$

\begin{eqnarray*}
 F_{\theta_1,\theta_2}((1-\alpha)\lambda_1+\alpha\lambda_2)&=& F\left(\theta_1+((1-\alpha)\lambda_1+\alpha\lambda_2)(\theta_2-\theta_1)\right),\\
&=& F( (1-\alpha)(\lambda_1(\theta_2-\theta_1)+\theta_1) + \alpha ((\lambda_2(\theta_2-\theta_1)+\theta_1)) ),\\
&<& (1-\alpha) F(\lambda_1(\theta_2-\theta_1)+\theta_1) + \alpha F((\lambda_2(\theta_2-\theta_1)+\theta_1)),\\
&<& (1-\alpha) F_{\theta_1,\theta_2}(\lambda_1) + \alpha  F_{\theta_1,\theta_2}(\lambda_2).
\end{eqnarray*}
\end{proof}

Then we define the multivariate Bregman  chord divergence by applying the definition of the univariate  Bregman  chord divergence of on these   families of univariate Bregman generators:

\begin{equation}\label{eq:CBDdD}
B_F^{\alpha,\beta}(\theta_1:\theta_2) \eqdef B_{F_{\theta_1,\theta_2}}^{\alpha,\beta}(0:1),
\end{equation}
Since $(01)_\alpha=\alpha$ and $(01)_\beta=\beta$, we get:
\begin{eqnarray*} 
&&B_F^{\alpha,\beta}(\theta_1:\theta_2)\\
&&= F_{\theta_1,\theta_2}(0)+\frac{\alpha (F_{\theta_1,\theta_2}(\alpha)-F_{\theta_1,\theta_2}(\beta))}{\beta-\alpha}-F_{\theta_1,\theta_2}(\alpha),\\
&& F(\theta_1) - F\left((\theta_1\theta_2)_\alpha\right) - \frac{\alpha \left( F\left((\theta_1\theta_2)_\beta\right) - F\left((\theta_1\theta_2)_\alpha\right)\right)}{\beta-\alpha},
\end{eqnarray*}
in accordance with the univariate case.
Since $(\theta_1\theta_2)_\beta=(\theta_1\theta_2)_\alpha-(\beta-\alpha)(\theta_2-\theta_1)$, we have the first-order Taylor expansion:
$$
F\left((\theta_1\theta_2)_\beta\right)\simeq_{\beta\simeq\alpha} F\left((\theta_1\theta_2)_\alpha\right)-(\beta-\alpha)(\theta_2-\theta_1)^\top\nabla F\left((\theta_1\theta_2)_\alpha\right).
$$
Therefore, we have:
$$\frac{\alpha \left( F\left((\theta_1\theta_2)_\beta\right) - F\left((\theta_1\theta_2)_\alpha\right)\right)}{\beta-\alpha}
\simeq -\alpha (\theta_2-\theta_1)^\top\nabla F\left((\theta_1\theta_2)_\alpha\right).
$$
This proves that $\lim_{\beta\rightarrow\alpha} B_F^{\alpha,\beta}(\theta_1:\theta_2)=B_F^{\alpha}(\theta_1:\theta_2)$.

Notice that the Bregman chord divergence does {\em not} require to compute the gradient $\nabla F$
The ``slope term'' in the definition is reminiscent to the $q$-derivative~\cite{qcalculus-2001} (quantum/discrete derivatives). 
However the $(p,q)$-derivatives~\cite{qcalculus-2001} are defined with respect to a single reference point while the chord definition requires two reference points.

\section{Conclusion and perspectives}\label{sec:concl}

We geometrically designed a new class of distances using two scalar parameters, termed the {\em Bregman chord divergence}, and its one-parametric subfamily, the {\em Bregman tangent divergences} that includes the ordinary Bregman divergence.
This generalization allows one to easily fine-tune Bregman divergences in applications by adjusting smoothly one or two (scalar) knobs. Moreover, by choosing $\alpha=1-\epsilon$ and   $\beta=1$ for small $\epsilon>0$, the Bregman chord divergence 
 $B_{F}^{1-\epsilon,1}(\theta_1:\theta_2)$
 lower bounds closely the Bregman divergence $B_{F}(\theta_1:\theta_2)$ without requiring to compute the gradient (a different approximation without gradient is $\frac{1}{\epsilon} J_F^\epsilon(\theta_2:\theta_1)$).
We expect that this new class of distances brings further improvements in signal processing and information 
fusion applications~\cite{FusionDistributions-2018} (e.g., by tuning $B_{F_\KL}^{\alpha,\beta}$ or $B_{F_\IS}^{\alpha,\beta}$). 
While the Bregman chord divergence defines an ordinate gap on the exterior of the epigraph, the Jensen chord divergence~\cite{chordgap-2018} defines the gap inside the epigraph of the generator.
In future work, the information-geometric structure induced by the Bregman chord divergences (curved)  shall be investigated from the viewpoint of gauge theory~\cite{gauge-2018} and in constrast with the dually flat structures of Bregman manifolds~\cite{IG-2016}.

Java\texttrademark{} Source code is available for reproducible research.\footnote{\url{https://franknielsen.github.io/~nielsen/BregmanChordDivergence/}}

\section*{Acknowledgments}
We express our thanks to Ga\"etan Hadjeres
(Sony CSL, Paris) for his careful proofreading and feedback.

\bibliographystyle{plain}
\bibliography{ChordBregmanDivergenceBIB}

\end{document}